\documentclass[runningheads]{llncs}

\usepackage[mobile]{accv}

\usepackage{accvabbrv}

\usepackage{graphicx}
\usepackage{booktabs}
\usepackage{caption}
\usepackage{subcaption}
\usepackage{multirow}

\usepackage[accsupp]{axessibility}  

\usepackage{amsmath,scalerel}

\usepackage[pagebackref,breaklinks,colorlinks,citecolor=accvblue]{hyperref}
\usepackage{hyperref}
\hypersetup{
    colorlinks=true,
    linkcolor=magenta,
    urlcolor=magenta
}
\usepackage{orcidlink}

\begin{document}

\title{VideoPatchCore: An Effective Method to Memorize Normality for Video Anomaly Detection} 

\titlerunning{VideoPatchCore}

\author{Sunghyun Ahn\inst{1}\orcidlink{0000-0001-9881-5731} \and
Youngwan Jo\inst{1}\orcidlink{0000-0003-2773-7670} \and
Kijung Lee\inst{1}\orcidlink{0000-1111-2222-3333} \and
Sanghyun Park\inst{1,}\thanks{Corresponding author: Sanghyun Park (\email{sanghyun@yonsei.ac.kr})\\17th Asian Conference on Computer Vision (ACCV 2024)}\orcidlink{0000-0002-5196-6193}}

\authorrunning{S. Ahn et al.}

\institute{Department of Computer Science, Yonsei University, Seoul, Republic of Korea \\
\email{\{skd, jyy1551, rlwjd4177, sanghyun\}@yonsei.ac.kr}}

\maketitle

\begin{abstract}
Video anomaly detection (VAD) is a crucial task in video analysis and surveillance within computer vision. Currently, VAD is gaining attention with memory techniques that store the features of normal frames. The stored features are utilized for frame reconstruction, identifying an abnormality when a significant difference exists between the reconstructed and input frames. However, this approach faces several challenges due to the simultaneous optimization required for both the memory and encoder-decoder model. These challenges include increased optimization difficulty, complexity of implementation, and performance variability depending on the memory size. To address these challenges, we propose an effective memory method for VAD, called VideoPatchCore. Inspired by PatchCore, our approach introduces a structure that prioritizes memory optimization and configures three types of memory tailored to the characteristics of video data. This method effectively addresses the limitations of existing memory-based methods, achieving good performance comparable to state-of-the-art methods. Furthermore, our method requires no training and is straightforward to implement, making VAD  tasks more accessible. Our code is available online at \href{https://github.com/SkiddieAhn/Paper-VideoPatchCore}{\textcolor{magenta}{github.com/SkiddieAhn/Paper-VideoPatchCore}}.
\end{abstract}

\begin{figure}
    \centering
    \begin{subfigure}[h]{1.0\linewidth}
        \centering
        \includegraphics[width=\textwidth]{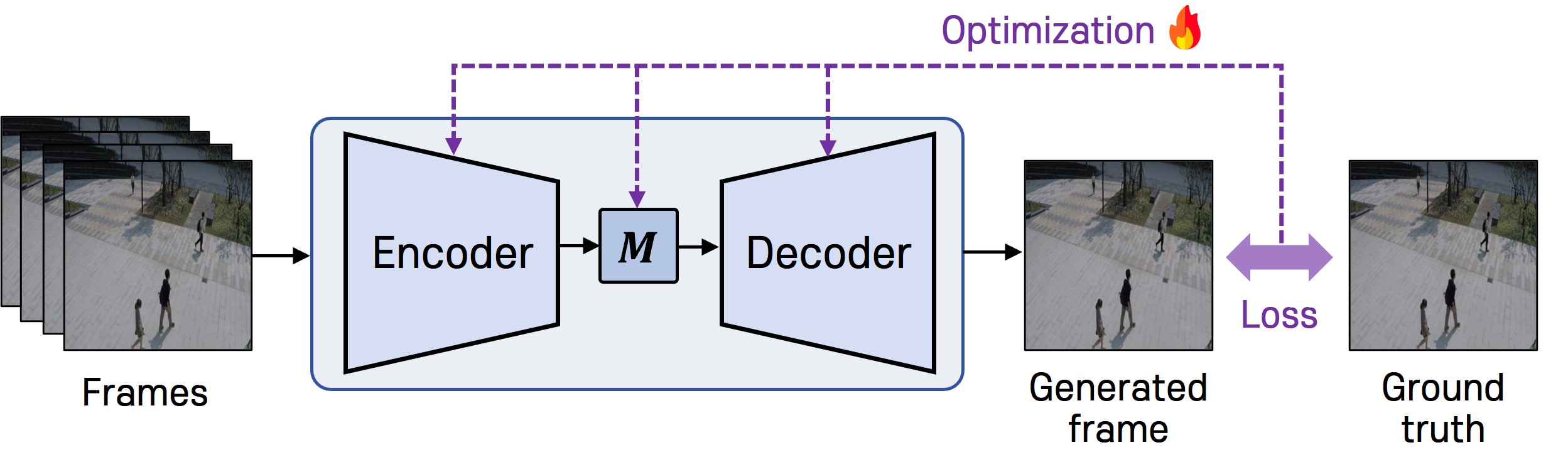}
        \caption{Memory-augmented Method}
        \label{fig:figure1a}
    \end{subfigure}
    \vspace{2pt}
    \vspace{2pt}
    \begin{subfigure}[h]{1.0\linewidth}
        \includegraphics[width=0.92\textwidth]{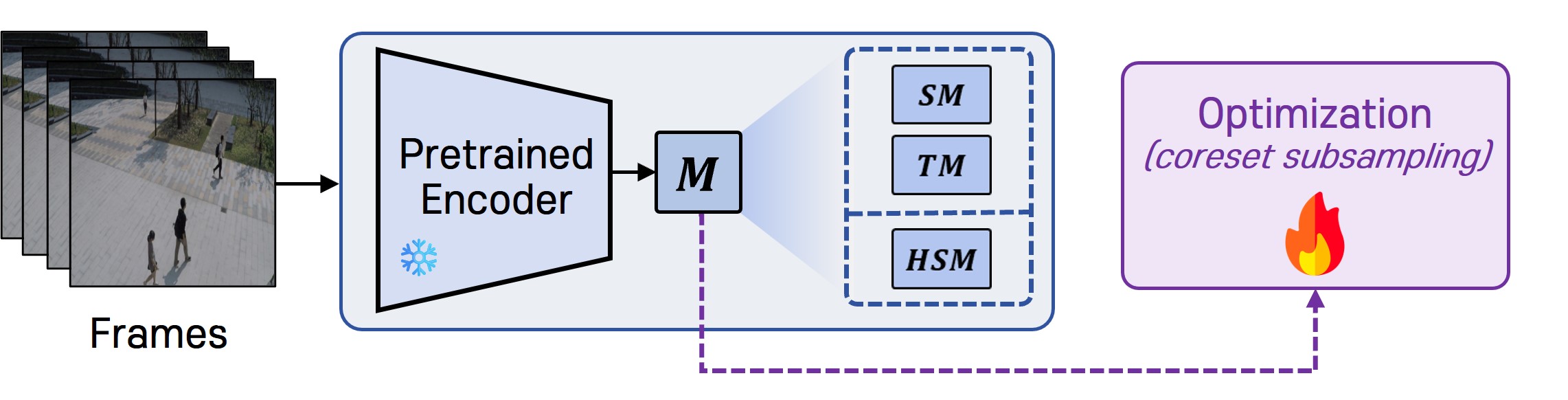}
        \caption{VideoPatchCore}
        \label{fig:figure1b}
    \end{subfigure}
    \caption{(a) Conventional memory-augmented methods optimize both the memory and the encoder-decoder model, simultaneously. (b) The proposed VideoPatchCore focuses on memory optimization.}
    \label{fig:figure1}
\end{figure}

\section{Introduction}
\label{sec:intro}
The task of video anomaly detection (VAD) involves identifying anomalies within video sequences. Typically, deep learning-based VAD methods employ one-class classification (OCC) techniques as anomalies are rare and it is challenging to define them precisely. OCC methods train models exclusively with normal data. In VAD, prominent OCC approaches include reconstruction and prediction-based methods \cite{hasan2016learning, nguyen2019anomaly, zaheer2020old, liu2018future, yang2023video}. These approaches utilize encoder-decoder models such as autoencoders (AEs), generative adversarial networks (GANs) \cite{goodfellow2014generative}, and others. During training, these models train to reconstruct or predict frame using only normal data, operating under the assumption that anomalies will result in high reconstruction or prediction errors during testing. However, these methods face challenges due to the exceptional generalization ability of deep learning models, which can generate plausible outputs for anomalous data \cite{zaheer2020old, huang2022self, huang2021abnormal, hong2024making}.

Recently, methods that store the features of normal data in memory during training have shown improved performance in reducing the generalization ability for anomalies. Memory-augmented VAD approaches \cite{gong2019memorizing, park2020learning, cai2021appearance, lv2021learning, liu2021hybrid, yang2022dynamic, sun2023learning, liu2023diversity, sun2023hierarchical} store the features of normal frames in memory and retrieve similar features from the memory to generate the input or future frame during testing. In such cases, anomalies result in higher reconstruction or prediction errors because their similar features are not stored in the memory. \cref{fig:figure1a} illustrates these memory utilization approaches. While memory-augmented methods demonstrate good performance, they also present several challenges. First, the simultaneous optimization of the model and memory makes the optimization process difficult and the implementation complex. Additionally, the size of the memory significantly impacts performance \cite{zhong2022cascade, Astrid2021LearningNT}. Although many recent studies \cite{park2020learning, sun2023hierarchical} have focused on reducing memory size, finding an appropriate memory size for large video datasets remains a challenge.

Inspired by PatchCore\cite{Roth2021TowardsTR}, we propose VideoPatchCore (VPC) to address the limitations of the previously mentioned memory utilization methods. As shown in \cref{fig:figure1b}, VPC leverages the vision encoder of the pre-trained CLIP \cite{Radford2021LearningTV} without additional training and optimizes memory using greedy coreset subsampling \cite{Agarwal2007GeometricAV}. This enables effective VAD with a small memory footprint even for large datasets. Unlike PatchCore, which is used at the image level, VPC performs anomaly detection at the video level by using two streams: local and global. The local stream detects anomalies in individual objects, while the global stream detects anomalies in entire frames. The local stream employs a spatial memory bank to identify the appearance anomalies and a temporal memory bank to detect action anomalies. The global stream utilizes a high-level semantic memory bank to detect anomalies in interactions among multiple objects and abnormalities that can only be identified by analyzing the scene (\eg, wrong direction). By utilizing spatial and temporal memory banks, VPC considers the spatiotemporal characteristics of videos, and with the two-stream approach, it can detect various forms of anomalies.

Through experiments, we demonstrate the effectiveness of using three memory banks in the video domain. Additionally, we introduce two methods for effectively memorizing video characteristics and demonstrate the robustness of performance concerning memory size on large datasets. Our proposed VPC shows good performance comparable to state-of-the-art techniques. We anticipate that VPC will enhance the accessibility of the VAD field and facilitate its use in real-world applications.

\noindent Our contributions are as follows:
\begin{itemize}
    \item[\textbullet] We propose VPC, an extension of PatchCore developed for image anomaly detection, to perform effective video anomaly detection.
    \item[\textbullet] VPC employs two streams (local and global) and three memory banks (spatial, temporal, and high-level semantic) to capture the spatiotemporal characteristics of videos and detect various forms of anomalies.
    \item[\textbullet] VPC achieves good performance comparable to state-of-the-art methods.
\end{itemize}

\section{Related Work}
\subsection{Reconstruction \& prediction-based Video Anomaly Detection}
The reconstruction-based approach is based on the assumption that a model trained to reconstruct normal video frames fails to accurately reconstruct abnormal frames. Nguyen \cite{nguyen2019anomaly} proposed a model comprising one encoder and two decoders. This model considered the $t^{th}$ frame as the input and predicted both the reconstructed frame and the optical flow between the $t^{th}$ and $t+1^{th}$ frames. Zaheer \cite{zaheer2020old} used GANs to generate high-quality reconstructed frames. In this approach, the discriminator was trained to classify high-quality frames as normal and low-quality frames as abnormal, utilizing a generator in its previous state that produced low-quality frames.

The prediction-based approach is based on the assumption that when a model is trained to predict future or past frames of normal video sequences, it fails to accurately predict abnormal frames. Liu \cite{liu2018future} utilized FlowNet and GANs to predict the $t+1^{th}$ frame from $t$ frames. Yang \cite{yang2023video} suggested selecting key frames from $t$ frames and predicting all $t$ frames based only on these key frames. 

\subsection{Video Anomaly Detection using Memory}
In VAD, memory is utilized to address the issue of accurately generating abnormal frames in OCC method. Gong 
\cite{gong2019memorizing} proposes a method that stores the features of normal frames in memory and generates normal frames using these stored memory items. Park \cite{park2020learning} suggests a method for learning various patterns of normal data by clustering similar features and using the cluster centers as memory items, thus storing numerous normal features in a compact memory. Sun \cite{sun2023hierarchical} employs contrastive learning to cluster features based on their classes and stores each feature in memory. This technique allows for the creation of a compact memory by leveraging the semantic 
information of objects. 

\subsection{Representation-based Image Anomaly Detection}
In the image domain, representation-based approaches have gained attention as effective 
methods for anomaly detection. This approach utilizes pre-trained networks on large-scale 
datasets such as ImageNet\cite{deng2009imagenet} and compares the embeddings of normal and abnormal images to detect anomalies. Cohen\cite{Cohen2020SubImageAD} proposes feature pyramid matching, leveraging the different information at each layer of the network. Features from the training data are stored in memory, and anomaly scores for the features of test data are computed using k-NN with respect to the features stored in memory. Roth\cite{Roth2021TowardsTR} points out that high-level features of pre-trained networks are specialized for classification purposes, indicating the need to extract features at the middle level. These features are stored in memory, and memory optimization is achieved through Coreset Subsampling.

\section{Method}
\subsection{Overview}
\begin{figure}
    \centering
    \includegraphics[width=\textwidth]{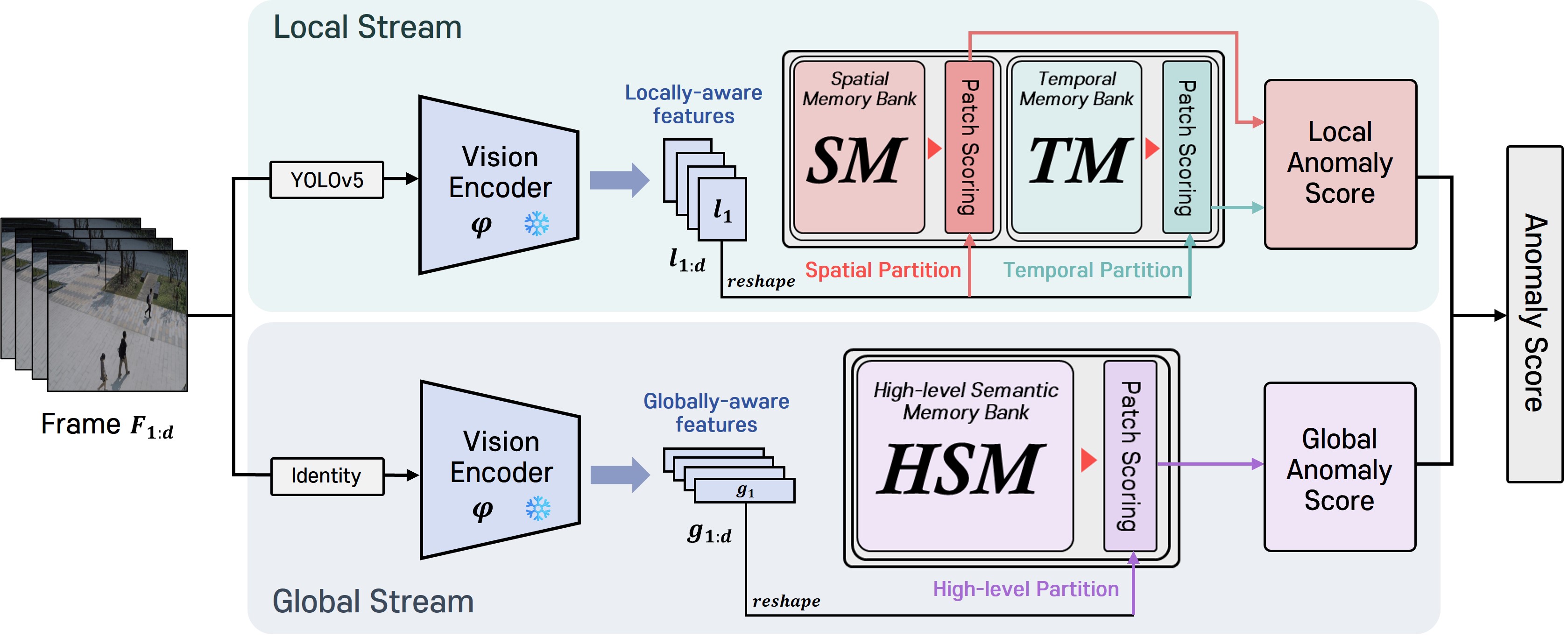}
    \caption{Architecture of VideoPatchCore (VPC). It consists of two streams (local and global) and three memory banks (spatial, temporal, and high-level semantic). The Spatial Memory Bank deals with the appearance information of objects to identify inappropriate objects. The Temporal Memory Bank handles the motion information of objects to detect inappropriate actions. The High-level Semantic Memory Bank processes the global context of frames to identify anomalies related to multiple objects or scenes.}
    \label{fig:figure2}
\end{figure}
\cref{fig:figure2} depicts the proposed VPC model, consisting of two streams: the local stream at the object level and the global stream at the frame level. Each stream extracts features through a vision encoder from CLIP, pre-trained on large-scale datasets, and these features are segmented according to the type of memory and accessed in the memory bank. During this process, the memorization and inference stages are performed. In the memorization stage, features are stored in memory, which is then optimized using greedy coreset subsampling from PatchCore. In the inference stage, anomaly scores are derived by calculating distances between memory and features. The following provides a detailed description of each component of the proposed VPC.\\ \\
\textbf{Vision Encoder.}
We adopt a CNN-based CLIP model as the Encoder and utilize layers 2, 3 $\varphi_2$, $\varphi_3$ similar to PatchCore. Typically, the shallow layers of the network capture the local features, while deeper layers capture the global features. However, the first layer captures overly detailed features such as edges and corners, and the last layer captures features biased towards classification, which are not helpful for anomaly detection. Thus, we process and utilize features from the intermediate layers $\varphi_2$, $\varphi_3$. \\ \\
\textbf{Local Stream.}
The local stream is composed of YOLOv5\cite{jocher2022ultralytics} for object detection, a vision encoder $\varphi$ for feature extraction, and spatial memory (SM) and temporal memory (TM) for storing the object appearance and motion information. Given a sequence of $d$ consecutive frames $F_1, F_2, \ldots, F_d \in \mathbb{R}^{3 \times H \times W}$, YOLOv5 generates a set of $n$ objects for each frame, $O_1, O_2, \ldots, O_d \in \mathbb{R}^{n \times 3 \times H \times W}$. Subsequently, the locally-aware features $l_1, l_2, \ldots, l_d \in \mathbb{R}^{n \times c \times h \times w}$are produced that represent the features of the objects.

Specifically, average pooling is performed on the features from $\varphi_2, \varphi_3$ and they are concatenated based on $\varphi_2$. This strategy aims to integrate both fine-grained and coarse-grained information that is necessary for object-level VAD. However, as $\varphi_2, \varphi_3$ have relatively small receptive fields, pooling is used to consolidate information over a wider range. The equation for constructing $l_i$ is as follows:
\begin{gather}
 l_i = \langle f_{\text{ap}}(\varphi_2(O_i)), f_{\text{ap}}(\varphi_3(O_i))\rangle \label{eqn:first} 
 \end{gather}
where $f_{\text{ap}}$ represents the average pooling and $\langle \cdot, \cdot \rangle$ denotes tensor concatenation.

Subsequently, $l_{1:d}$ are reshaped into $lf \in \mathbb{R}^{n \times c \times d \times h \times w}$. However, the dimensionality of $lf$ is impractically large for video processing. Therefore, we employ a method called split pooling to reduce the number of channels, transforming it into $\mathbb{R}^{n \times c' \times d \times h \times w}$ ($c \gg c'$). The details of this approach can be found in \Cref{sec:ablation_study}. Subsequently, $lf$ is split into spatial patches and temporal patches through spatial partition and temporal partition, respectively, and stored in SM and TM. \\ \\
\textbf{Global Stream.}
The global stream constructs globally-aware features $g_1, g_2, \\
\ldots, g_d \in \mathbb{R}^{c \times 1 \times 1}$ to capture the overall context of frames, thereby differing from the local stream, which focuses on individual objects within frames. We utilize $\varphi_2, \varphi_3$, applying global pooling to obtain global information. Specifically, average pooling is applied to the features of each layer, and these are concatenated based on $\varphi_3$, which handles more global information. During global pooling, both average and max pooling techniques are employed to achieve a better representation. The equation for constructing $g_i$ is as follows:
\begin{gather}
 g_i = \langle f_{\text{ap}}(\varphi_2(F_i)), f_{\text{ap}}(\varphi_3(F_i))\rangle \label{eqn:second} \\
 g_i = f_{\text{gap}}(g_i) + f_{\text{gmp}}(g_i) \label{eqn:third}
 \end{gather}
where $f_{\text{gap}}$ and  $f_{\text{gmp}}$ denote global average pooling and global max pooling, respectively.

Subsequently, $g_{1:d}$ are reshaped into $gf \in \mathbb{R}^{c \times d \times 1 \times 1}$. The $gf$ constructed in this manner includes the global information of each frame and is converted into high-level patches through high-level partition. Afterwards, the patch is stored in the High-level Semantic Memory (HSM).

\subsection{Patch Partition}
\begin{figure}
    \centering
    \includegraphics[width=\textwidth]{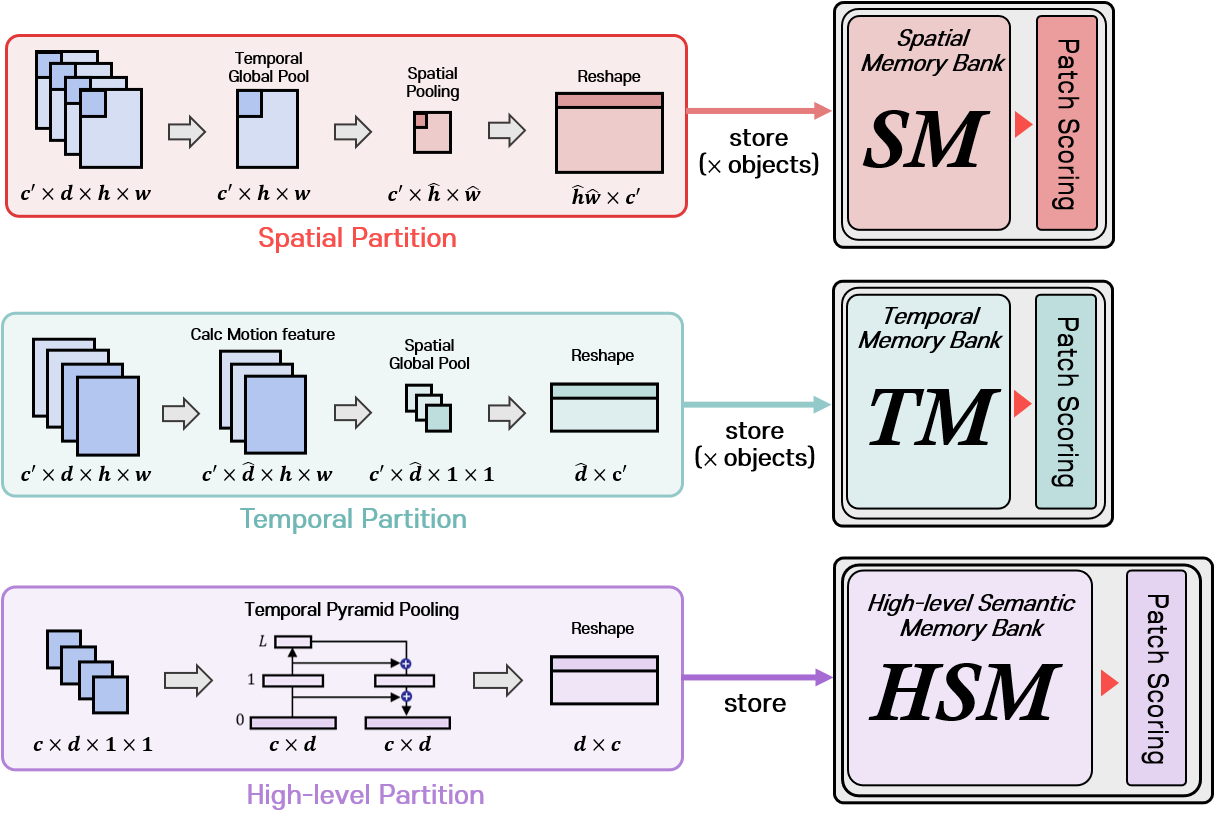}
    \caption{Patch Partition Method. `objects' denotes the number of objects(n), and objects are processed in parallel considering batch processing.}
    \label{fig:figure3}
\end{figure}

\cref{fig:figure3} illustrates the proposed patch partition method, which consists of spatial partition, temporal partition, and high-level partition. The spatial partition focuses on object appearance information, generating patches while disregarding temporal information. The temporal partition emphasizes object motion information, generating patches while ignoring spatial information. The high-level partition utilizes extensive spatiotemporal information to generate patches for understanding the contextual information across frames. The detailed descriptions of each partition are as follows.\\ \\
\textbf{Spatial Partition.}
Appearance information is crucial for evaluating anomalies in objects and can be derived from spatial features. We compress the input feature map $lf$ using temporal global pooling\cite{kwon2018video} to preserve spatial information while ignoring temporal aspects. Subsequently, average pooling is applied to consider various regions of the object. For instance, when assessing a person's pose, it is essential to consider multiple joints together. The equation for creating $Spatial Patches$ is as follows:
\begin{gather}
Spatial Patches = f_{\text{ap}}(f_{\text{tgp}}(lf)) \label{eqn:fourth}
\end{gather} 
where $f_{\text{tgp}}$ denotes the temporal global pooling. Finally, the results are reshaped into $\mathbb{R}^{(n \cdot \hat{h} \cdot \hat{w}) \times c'}$, where $h \geq \hat{h}$ and $w \geq \hat{w}$.\\ \\
\textbf{Temporal Partition.}
Motion information in objects represents changes over time, making it crucial for VAD. We utilize the adjacent temporal information within $lf$ to generate motion features $mf$. To achieve this, the feature differences are computed and the representative motion values are determined through global average pooling. The equation for creating $Temporal Patches$ is as follows:
\begin{gather}
mf_{(t)} = |lf_{(t+1)} - lf_{(t)}| \label{eqn:fifth}\\
Temporal Patches = f_{\text{gap}}(mf) \label{eqn:sixth}
\end{gather} 
 \( mf_{(t)}\in\mathbb{R}^{n \times c' \times h \times w} \) represents the difference between the $t^{th}$ and $t+1^{th}$ time within $lf$, and it belongs to \( mf \in \mathbb{R}^{n \times c' \times \hat{d} \times h \times w} \), where $\hat{d}=d-1$. Finally, the results are reshaped into $\mathbb{R}^{(n \cdot \hat{d}) \times c'}$.\\ \\
\textbf{High-level Partition.}
In frames, the global context processes the relationship between the objects and the scene, and considers interactions between different objects, thereby enabling more accurate VAD. As $gf$ already encompasses the global context from a spatial perspective, temporal pyramid pooling is utilized to obtain high-level temporal information. Additionally, this approach secures multi-scale temporal information, addressing the limitation of only using adjacent temporal information in temporal partition methods. The equation for creating $Highlevel Patches$ is as follows:
\begin{gather}
Highlevel Patches = \sum_{l=0}^{L}  f_{\text{mp}}^l (gf) \label{eqn:seventh}
\end{gather} 
Temporal pyramid pooling is implemented using $f_{\text{mp}}^l$. In this case, $f_{\text{mp}}^l$ represents applying the max pooling operation $l$ times. Finally, the results are reshaped into $\mathbb{R}^{d \times c}$. 
\subsection{Inference}
Similar to the memorization stage, the Patch Partition method is applied as described earlier. At this time, mismatched pooling technology is used to improve the VAD performance. This technique and the method of calculating the anomaly score are as follows. \\ \\
\textbf{Mismatched Pooling.}
During the temporal partition process, global pooling is applied differently in the memorization and inference stages. The memorization stage employs global average pooling (GAP), while global max pooling (GMP) is used during inference. As normal objects typically exhibit more static characteristics compared to abnormal ones, we expected normal motion features to show similar average and maximum values. In contrast, abnormal motion features were anticipated to show significant discrepancies in these values. By adopting this approach, VAD performance improves by preventing the retrieval of nearby temporal patches from memory when abnormal data is input. Experimental validation of the effectiveness of this methodology is detailed in \Cref{sec:ablation_study}. \\ \\
\textbf{Anomaly Scoring.}
All nearest neighbors are computed between patches $X$ generated from test frames and the memory bank $M$. Subsequently, the representative patch $x^*$ is selected from the $X$ and its nearest neighbor $m^*$ is selected from the $M$. The distance $s^*$ between $x^*$ and $m^*$ is assigned as the score representing the frames.
\begin{gather}
x^*, m^* = \underset{x \in X}{\mathrm{arg\,max}} \, \underset{m \in M}{\mathrm{arg\,min}} \, \|x - m\|_2 \label{eqn:eighth} \\
s^* = \|x^* - m^*\|_2  \label{eqn:nineth} 
\end{gather} 
As three types of memory are used, $s^*$ is computed three times for SM, TM, and HSM, respectively. The $s_{\text{spatial}}^*$ calculated from SM and $s_{\text{temporal}}^*$ calculated from TM determine the local anomaly score (LAS) in the local stream. Conversely, $s_{\text{high-level}}^*$ computed from HSM is utilized as the global anomaly score (GAS). Ultimately, the LAS and GAS combine to derive the anomaly score. The equation for calculating the anomaly score is as follows:
\begin{gather}
LAS = \delta_1 \cdot s_{\text{spatial}}^* + \delta_2 \cdot s_{\text{temporal}}^* \label{eqn:tenth} \\
GAS = s_{\text{high-level}}^* \label{eqn:eleventh} \\
Anomaly Score = \gamma_1 \cdot \frac{LAS - \mu(LAS)}{\sigma(LAS)} + \gamma_2 \cdot \frac{GAS - \mu(GAS)}{\sigma(GAS)} \label{eqn:twelveth}
\end{gather} 

\section{Experiments}
\subsection{Datasets}
The proposed model was validated on three datasets: CUHK Avenue (Avenue)\cite{lu2013abnormal}, ShanghaiTech (SHTech)\cite{zhang2016single}, and IITB Corridor (Corridor)\cite{rodrigues2020multi}. As each dataset was constructed for OCC method, the training dataset consisted only of normal videos, while the test dataset contained both normal and abnormal videos. \\ \\
\textbf{Avenue.}
The dataset consisted of 16 training and 21 testing videos filmed with a single camera at the same angle. It included anomaly events such as throwing paper, running on the road, and walking in the wrong direction.\\ \\
\textbf{SHTech.}
The dataset consisted of 330 training videos and 107 testing videos, capturing 13 scenes  from different angles. It included various anomaly events such as the appearance of anomaly objects (\eg, cars, bicycles) and action anomalies (\eg, fighting, jumping).\\ \\
\textbf{Corridor.}
The dataset consisted of 208 training and 150 testing videos filmed with a single camera at the same angle. It included previously unseen anomaly events such as protesting, hiding, and playing with the ball.

\subsection{Implementation details}
\textbf{Object Detection.}
The pretrained YOLOv5 was utilized for object detection in the videos. The bounding boxes of the objects were generated based on the last or middle frames among the $d$ input frames provided, and these bounding boxes were shared across other frames. Specifically, for the Avenue, SHTech, and Corridor datasets, we used 10, 4, 4 input frames, respectively. For the Avenue dataset, objects were detected based on the middle frame. Furthermore, to minimize the loss of temporal information, margins were added to the width and height of the bounding boxes, ensuring that they remained equal in size. Finally, the object images were resized to (224, 224) before being input into the pre-trained network. \\ \\
\textbf{Memorizing details.}
We utilized CLIP's ResNet-101 as the vision encoder. The value $c'$ for split pooling in the local stream was  determined by the dataset size. Specifically, it was set to 32 for the smaller Avenue dataset and 64 for the larger SHTech and Corridor datasets. For $lf$ and $gf$ generation, average pooling with a kernel size of $(3, 3)$ and stride of 1 was used. In the spatial partition, average pooling was performed with a kernel size of $(4, 4)$. In the high-level partition, $L$ in TPP was set to 2, and during max pooling, a kernel size of $(2)$ with a stride of 1 was used. During memory optimization, coreset subsampling rates were set to 1, 25, and 10\% for the Avenue, SHTech, and Corridor datasets, respectively.
\begin{table}[t]
\centering
\caption{Comparison of VPC against SOTA methods in terms of AUROC. Avenue: CUHK Avenue, SHTech: ShanghaiTech, Corridor: IITB Corridor. The best results are \textbf{bolded}. The second-best results are \underline{underlined}. *experimented by us.}
\label{tab:auc_comparison}{
\begin{tabular}{cccccc}
\toprule
\textbf{Method} & \textbf{Venue} & \textbf{Memory} & \textbf{ Avenue } & \textbf{SHTech } & \textbf{Corridor} \\
\addlinespace[2pt]
\hline
FFP\cite{liu2018future} & CVPR 18 &  &  84.9\%  & 72.8\% & 64.7\%     \\
\hline
MPED-RNN\cite{morais2019learning} & CVPR 19 &  &  -  & 73.4\% & 64.3\%     \\
MemAE\cite{gong2019memorizing} & ICCV 19 & \checkmark &  83.3\%  & 71.2\% & -     \\
AMC\cite{nguyen2019anomaly} & ICCV 19 &  &  86.9\%  & - & -     \\
\hline
MTP\cite{rodrigues2020multi} & WACV 20 &  &  82.9\%  & 76.0\% & 67.1\%     \\
CDDA\cite{chang2020clustering} & ECCV 20 &  &  86.0\%  & 73.3\% &  -     \\
MNAD\cite{park2020learning} & CVPR 20 & \checkmark &  88.5\%  & 70.5\% &  -     \\
\hline
ROADMAP\cite{wang2021robust} & TNNLS 21 &  &  88.3\%  & 76.6\% & -     \\
AMMC-Net\cite{cai2021appearance} & AAAI 21 & \checkmark &  86.6\%  & 73.7\% & -     \\
MPN\cite{lv2021learning} & CVPR 21 & \checkmark &  89.5\%  & 73.8\% & -     \\
$\text{HF}^{2}\text{-VAD}$\cite{liu2021hybrid} & ICCV 21 & \checkmark &  91.1\%  & 76.2\% & -     \\
\hline
LLSH\cite{lu2022learnable} & TCSVT 22 &  &  87.4\%  & 77.6\% & 73.5\%     \\
VABD\cite{li2021variational} & TIP 22 &  &  86.6\%  & 78.2\% & 72.2\%     \\
DLAN-AC\cite{yang2022dynamic} & ECCV 22 & \checkmark  &  89.9\%  & 74.7\% & -     \\
Jigsaw\cite{wang2022video} & ECCV 22 &  &  92.2\%  & 84.3\% & -     \\
\hline
Sun et al.\cite{sun2023learning} & AAAI 23 & \checkmark &  91.5\%  & 78.6\% & -     \\
Cao et al.\cite{cao2023new} & CVPR 23 &  &  86.8\%  & 79.2\% & \underline{73.6\%}     \\
USTN-DSC\cite{yang2023video} & CVPR 23 &  &  89.9\%  & 73.8\% & -     \\
DMAD\cite{liu2023diversity} & CVPR 23 & \checkmark &  \textbf{92.8\%}  & 78.8\% & -     \\
FPDM\cite{yan2023feature} & ICCV 23 &  &  90.1\%  & 78.6\% & -     \\
$\text{STG-NF}^{*}$\cite{hirschorn2023normalizing} & ICCV 23 &  &  61.8\% & \textbf{85.9\%} & 61.4\%     \\
HSC\cite{sun2023hierarchical} & ICCV 23 & \checkmark &  \underline{92.4\%}  & 83.0\% & -     \\
\hline
Ristea et al.\cite{ristea2024self} & CVPR 24 &  &  91.3\%  & 79.1\% & -     \\
Zhang et al.\cite{zhang2024multi} & CVPR 24 &  &  \underline{92.4\%}  & \underline{85.1\%} & -     \\
\hline
\addlinespace[2pt]
\textbf{VPC (Ours)} & - & \checkmark &  \textbf{92.8\%}  & \underline{85.1\%} & \textbf{76.4\%}     \\
\bottomrule
\end{tabular}
}
\end{table}
\\ \\
\textbf{Inference details.}
The anomaly score was smoothed using a 1D Gaussian filter during testing. The parameters $\delta_1$ and $\delta_2$ used in LAS computation were set to $(0.7, 0.3)$ for Avenue and $(0.5, 0.5)$ for the SHTech and Corridor datasets. The parameters $\gamma_1$ and $\gamma_2$ used in the final anomaly score computation were set to $(0.7, 0.3)$ for Avenue and $(0.9, 0.1)$ for the SHTech and Corridor datasets. The experiments were conducted using an NVIDIA GeForce RTX 3090 graphics card.

\subsection{Evaluation Criteria}
Consistent with prior research, the frame-level area under the curve (AUC) of Receiver Operation Characteristic (ROC) was used to evaluate the performance of the proposed model. This evaluation was calculated as the area under the ROC curve by varying the threshold. Specifically, we aggregated anomaly scores across all frames in the dataset to calculate the micro-averaged AUROC.

\subsection{Comparison with SOTA models}
\cref{tab:auc_comparison} presents the results of the proposed model's performance on three benchmark datasets. It demonstrates the competitive performance compared to other state-of-the-art (SOTA) methods. Specifically, when compared to HSC\cite{sun2023hierarchical}, which utilizes appearance and motion memory, our approach outperforms by 0.4\% and 2.1\% on the Avenue and SHTech datasets, respectively, leveraging three memory components effectively. And, compared to FPDM\cite{yan2023feature}, which also employs the CLIP, our model achieves better performance by accurately detecting anomaly events at the local stream. Furthermore, compared to the other methods using memory, a superior performance is achieved. This demonstrates that a method focused on memory optimization is more effective than traditional methods using memory. In addition, individuals without deep learning expertise can easily use the proposed approach because it does not require training.

\begin{figure}[!b]
    \centering
    \begin{subfigure}[h]{1.0\linewidth}
        \centering
        \includegraphics[width=\textwidth]{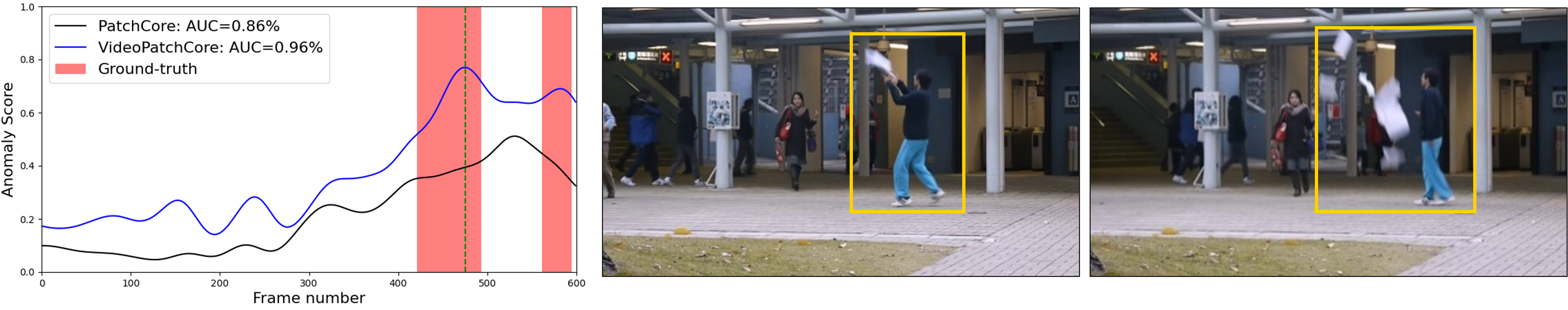}
        \caption{A person throwing paper}
        \label{fig:figure4a}
    \end{subfigure}
    \begin{subfigure}[h]{1.0\linewidth}
        \includegraphics[width=\textwidth]{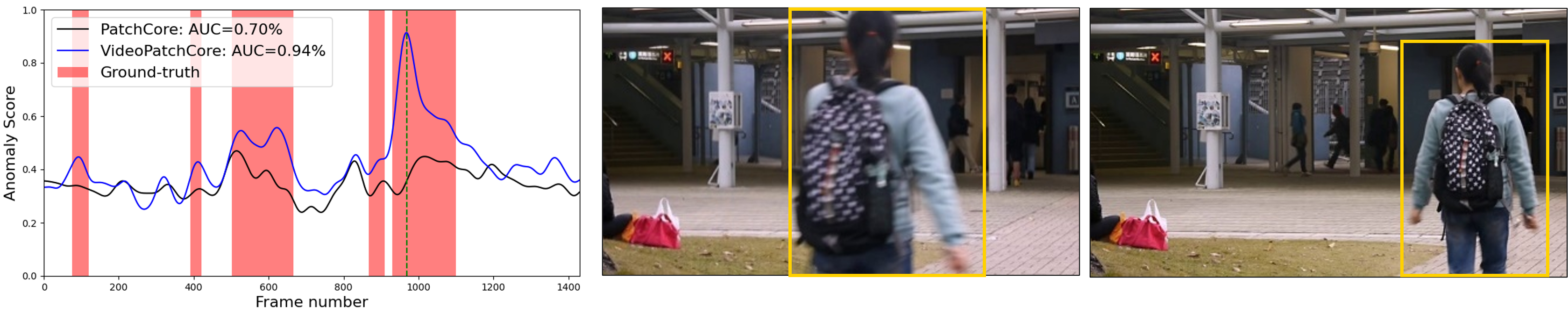}
        \caption{Wrong direction}
        \label{fig:figure4b}
    \end{subfigure}
    \caption{Comparison of the anomaly scores between VideoPatchCore and PatchCore in the Avenue dataset.}
    \label{fig:figure4}
\end{figure}
\subsection{Qualitative Evaluation}
To  intuitively understand the use of multiple memories, for the Avenue dataset, we compared PatchCore (PC), which utilizes appearance information, with \\VideoPatchCore (VPC), which incorporates not only appearance information, but also motion and high-level information. Here, PC refers to the use of spatial memory only in VPC. \cref{fig:figure4a} depicts an anomalous frame showing the action of throwing papers, necessitating consideration of motion information. Therefore, temporal memory plays a crucial role in this scenario. \cref{fig:figure4b} depicts an anomalous frame showing the action of walking in the incorrect direction, necessitating consideration of both object and scene contexts. Therefore, high-level semantic memory plays a crucial role in this scenario. It is known that VPC, which utilizes temporal and high-level semantic memory, detects anomalies well, whereas PC, which does not utilize them, does not detect anomaly. Additional experiments on the SHTech and Corridor dataset are presented in the supplementary materials.

\subsection{Ablation study}
\label{sec:ablation_study}
\begin{table}[tp]
\centering
\caption{Comparison of the AUROC scores for the spatial and temporal memory on the Avenue, SHTech 
and Corridor datasets.}
\label{tab:spatial_temporal_auc}
\begin{tabular}{cccc}
\toprule
\textbf{Memory} & \textbf{\quad Avenue \quad} & \textbf{\quad SHTech \quad} & \textbf{\quad Corridor \quad}\\
\hline\hline
\addlinespace[2pt]
Spatial & 84.8\% & 74.7\% & 70.5\% \\
Temporal & 66.9\% & 78.8\% & 73.5\%\\
Spatial+Temporal & \textbf{90.3\%} & \textbf{84.8\%} & \textbf{76.3\%}\\
\bottomrule
\end{tabular}
\end{table}

\begin{table}[tp]
\centering
\caption{Comparison of the AUROC scores for compression methods on the Avenue, SHTech 
and Corridor datasets.}
\label{tab:split_pool_auc}
\begin{tabular}{cccc}
\toprule
\textbf{Compression} & \textbf{\quad Avenue \quad} & \textbf{\quad SHTech \quad} & \textbf{\quad Corridor \quad}\\
\hline\hline
\addlinespace[2pt]
Head & 82.7\% & 76.9\% & 74.9\% \\
Random & 85.9\% & 83.8\% & 74.8\%\\
Split Pool & \textbf{90.3\%} & \textbf{84.8\%} & \textbf{76.3\%}\\
\bottomrule
\end{tabular}
\end{table}

\textbf{Spatial and Temporal memory.}
\cref{tab:spatial_temporal_auc} presents a comparison of the VAD performance affected by the spatial and temporal memory. The spatial memory effectively detects the appearance of the anomaly object, while temporal memory effectively detects the action anomaly event. As it is important to consider both the motion and appearance information for accurate anomaly detection, using both types of memory helps achieve high performance on the Avenue, SHTech, and Corridor datasets.\\ \\
\textbf{Split pooling.}
\cref{tab:split_pool_auc} presents the AUROC scores based on information compression techniques in the local stream. Head compresses $lf$ by selecting $c'$ channels from the front. Random compresses $lf$ by selecting random $c'$ channels. Split pool divides $lf$ into $c'$ groups and compresses by averaging the channels of each group. The common compression methods lose performance as they fail to preserve the original information, whereas the proposed compression method achieves high performance by effectively compressing the original information.\\ \\
\textbf{Mismatched pooling.}
\cref{tab:pooling_auc} presents the AUROC scores based on feature pooling methods in the temporal partition. The avg and max pool perform GAP and GMP during the memorization and inference stage, respectively. The common pooling methods fail to effectively detect action anomalies, whereas the proposed method achieves effective detection. This demonstrates that differentiating between normal and abnormal temporal patches by varying the pooling methods at each stage is effective for VAD.
\begin{table}[tp]
\centering
\caption{Comparison of AUROC scores for feature pooling methods in the temporal partition across the Avenue, SHTech 
and Corridor datasets.}
\label{tab:pooling_auc}
\begin{tabular}{cccc}
\toprule
\textbf{Pooling} & \textbf{\quad Avenue \quad} & \textbf{\quad SHTech \quad} & \textbf{\quad Corridor \quad}\\
\hline\hline
\addlinespace[2pt]
Avg Pool & 84.9\% & 76.1\% & 74.7\% \\
Max Pool & 88.2\% & 83.8\% & 75.7\%\\
Mismatched Pool & \textbf{90.3\%} & \textbf{84.8\%} & \textbf{76.3\%}\\
\bottomrule
\end{tabular}
\end{table}
\begin{table}[tp]
\centering
\caption{Comparison of the AUROC scores for the local and global stream on the Avenue, SHTech and 
Corridor datasets.}
\label{tab:total_auc}
\begin{tabular}{cccc}
\toprule
\textbf{Stream} & \textbf{\quad Avenue \quad} & \textbf{\quad SHTech \quad} & \textbf{\quad Corridor \quad}\\
\hline\hline
\addlinespace[2pt]
Local & 90.3\% & 84.8\% & 76.3\% \\
Global & 84.4\% & 68.4\% & 67.2\%\\
Local+Global & \textbf{92.8\%} & \textbf{85.1\%} & \textbf{76.4\%}\\
\bottomrule
\end{tabular}
\end{table}
\\ \\
\textbf{Local and Global stream.}
\cref{tab:total_auc} presents a comparison of the anomaly detection performance affected by the local and global stream. When using only the local stream, it is difficult to detect anomalies based on scenes and abnormal behaviors among multiple objects. To address this issue, we incorporated the global stream, which utilizes a wide range of spatial-temporal information. When adding the global stream to Avenue, there is a significant performance improvement, whereas adding the global stream to SHTech and Corridor results in relatively slight performance improvements. In the latter datasets, there are many situations where objects are adjacent to each other, allowing the local stream alone to partially fulfill the role of the global stream.

\subsection{Further Analysis}
\textbf{Coreset subsampling ratio.}
\cref{tab:subsampling_ratio} presents the AUROC scores based on the subsampling ratio. 
Generally, increasing the sampling ratio in large datasets stores a diverse range of normal features in memory, leading to improved performance, but it can also result in slower processing speed. However, using Coreset subsampling, the performance difference in memory usage between 1\% and 99\% is very minimal, less than 0.5\%, and the frames per second (FPS) can be up to 7.3 times faster. This demonstrates the effectiveness of coreset subsampling in large datasets such as SHTech and Corridor.\\ \\
\textbf{Memorizing Technique.}
\cref{tab:memorizing_technique} presents the AUROC scores based on the subsampling order. The former method constructs memory from each video and concatenates them, whereas the latter method constructs memory from all videos collectively. We expected that the former method, which optimizes memory for each video, would be less constrained by hardware limitations but might suffer from a lower performance due to the uneven distribution of features. However, in fact, the performance difference between the two methods is almost nonexistent. This suggests that it is possible to sufficiently store a diverse range of normal features regardless of the order of subsampling.
\begin{table*}[t!]
\caption{Comparison of AUROC scores for the subsampling ratio on the SHTech and Corridor datasets.}
\label{tab:subsampling_ratio}
\centering{%
\begin{tabular}{cccccc}
\toprule
    \multirow{2}{*}{\textbf{Subsampling ratio}} &
    \multicolumn{2}{c}{\textbf{SHTech}} &
    \multicolumn{2}{c}{\textbf{Corridor}} &
    \\ \cmidrule(l){2-3} \cmidrule(l){4-5} 
    & \textbf{AUC} & \textbf{FPS} & \textbf{AUC} & \textbf{FPS} &  \\ \midrule
    \midrule

    \multicolumn{1}{c}{
    \begin{tabular}[c]{@{}l@{}}
        1\% \\
        10\% \\
        25\% \\
        50\% \\
        75\% \\
        99\% \\
    \end{tabular}} &
    
    \multicolumn{1}{c}{
    \begin{tabular}[c]{@{}c@{}} 
        84.6\% (-0.5\%)  \\
        85.0\% (-0.1\%)  \\
        \textbf{85.1\% (-0.0\%)} \\
        \textbf{85.1\% (-0.0\%)}  \\
        \textbf{85.1\% (-0.0\%)}  \\
        \textbf{85.1\%} 
    \end{tabular}} &
    \multicolumn{1}{c}{ 
    \begin{tabular}[c]{@{}c@{}} 
        \textbf{170.9} \\
        154.8 \\
        96.1 \\
        54.8 \\
        39.6 \\
        31.0 
    \end{tabular}} &

    \multicolumn{1}{c}{
    \begin{tabular}[c]{@{}c@{}} 
        76.0\% (-0.4\%)  \\
        \textbf{76.4\% (-0.0\%)} \\
        76.3\% (-0.1\%) \\
        76.3\% (-0.1\%) \\
        76.3\% (-0.1\%) \\
        \textbf{76.4\%}
    \end{tabular}} &
    \multicolumn{1}{c}{ 
    \begin{tabular}[c]{@{}c@{}} 
        \textbf{143.4} \\
        113.5 \\
        60.0 \\
        40.0 \\
        25.2 \\
        19.5
    \end{tabular}} &
    \\  \bottomrule
\end{tabular}
}
\end{table*}

\begin{table}[tp]
\centering
\caption{Comparison of AUROC scores for the memorizing technique at memory usage levels of both 10\% and 99\% on the SHTech and Corridor datasets. }
\label{tab:memorizing_technique}
\begin{tabular}{cccc}
\toprule
\textbf{Memorizing technique } & \textbf{\quad Ratio \quad} & \textbf{\quad SHTech \quad} & \textbf{\quad Corridor \quad}\\
\hline\hline
\addlinespace[2pt]
Subsampling  & 10\% & 85.0\% & \textbf{76.4\%} \\
$\rightarrow \text{concat}$ & 99\% & \textbf{85.1\%} & \textbf{76.4\%}\\
\midrule
Concat  & 10\% & 84.9\% (-0.1\%) & 76.3\% (-0.1\%) \\
$\rightarrow \text{subsampling}$ & 99\% & 85.0\% (-0.1\%) & \textbf{76.4\% (-0.0\%)}\\
\bottomrule
\end{tabular}
\end{table}
\section{Conclusion}
We present a VPC that extends PatchCore to the video level. The proposed model consists of three 
types of memory considering the characteristics of the video, and maintains high AUC and FPS through 
memory optimization using coreset subsampling. This approach can be used as an effective 
memory in VAD by solving three problems of traditional memory methods (increased optimization 
difficulty, complexity of implementation, and performance variability depending on the memory size). 
In addition, VPC shows excellent performance in all benchmark datasets and does not require 
training, increasing accessibility in the VAD field. We expect VPC to make a significant contribution 
to the VAD field.\\ \\
\textbf{Acknowledgements.}
This research was supported by the National Research Foundation (NRF) funded by the Korean government (MSIT) (No. RS-2023-00229822). The funders did not play any role in the design of the study, data collection, analysis, or preparation of the manuscript.

\bibliographystyle{splncs04}
\bibliography{VPC}

\newpage
\begin{center}
    \fontsize{14}{14}\selectfont \textbf{VideoPatchCore: An Effective Method to Memorize Normality for Video Anomaly Detection \emph{(Supplementary Materials)}}
\end{center}

\setcounter{section}{0}
\renewcommand\thesection{\Alph{section}}
\setcounter{figure}{0}
\renewcommand\thefigure{S\arabic{figure}}
\setcounter{table}{0}
\renewcommand\thetable{S\arabic{table}}

\section{Qualitative Evaluation}
\subsection{Anomaly Score Visualization}
To intuitively understand the use of multiple memories, we compared PatchCore (PC), which utilizes appearance information, with VideoPatchCore (VPC), which incorporates not only appearance information, but also motion and high-level information, on the SHTech and Corridor datasets.

\begin{figure}
    \centering
    \begin{subfigure}[h]{1.0\linewidth}
        \centering
        \includegraphics[width=\textwidth]{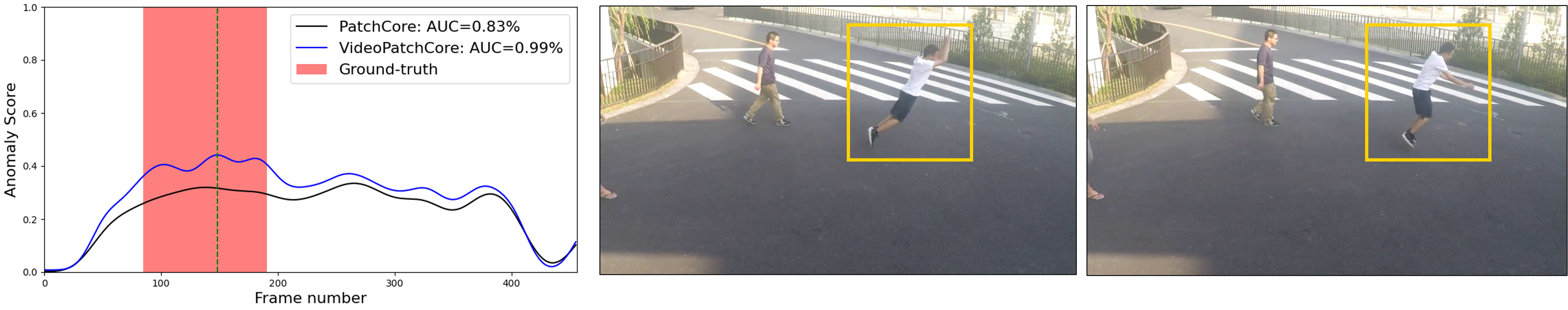}
        \caption{Jumping}
        \label{fig:sfigure1a}
    \end{subfigure}
    \begin{subfigure}[h]{1.0\linewidth}
        \includegraphics[width=\textwidth]{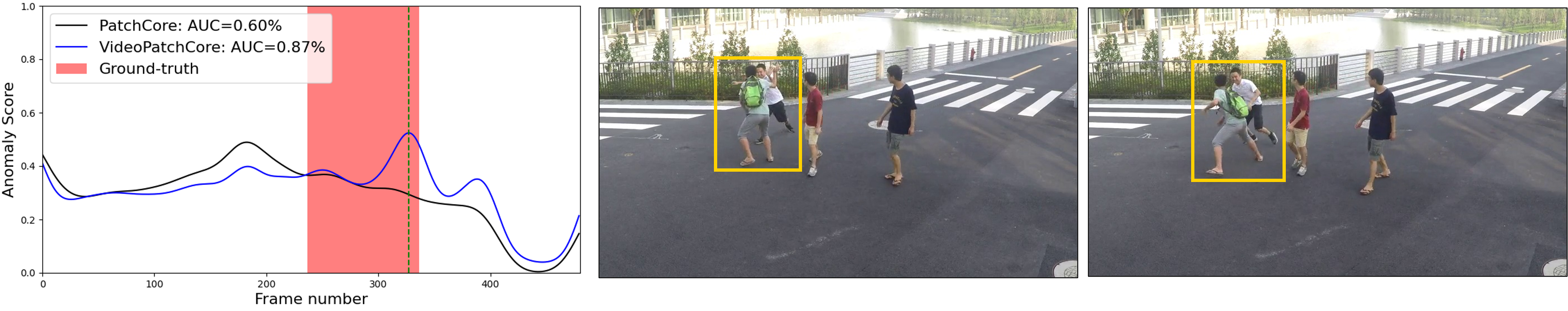}
        \caption{Fighting}
        \label{fig:sfigure1b}
    \end{subfigure}
    \caption{Comparison of anomaly scores between VPC and PC in the SHTech dataset.}
    \label{fig:sfigure1}
\end{figure}
\cref{fig:sfigure1a} shows anomaly frames depicting the action of jumping, necessitating consideration of motion information. Therefore, temporal memory plays a crucial role in this scenario. \cref{fig:sfigure1b} shows anomaly frames depicting the action of fighting, necessitating consideration of interactions between two peoples. Therefore, high-level semantic memory plays a crucial role in this scenario. It is known that VPC, which utilizes temporal and high-level semantic memory, detects anomalies well, whereas PC, which does not utilize them, does not detect anomalies.

\begin{figure}
    \centering
    \begin{subfigure}[h]{1.0\linewidth}
        \centering
        \includegraphics[width=\textwidth]{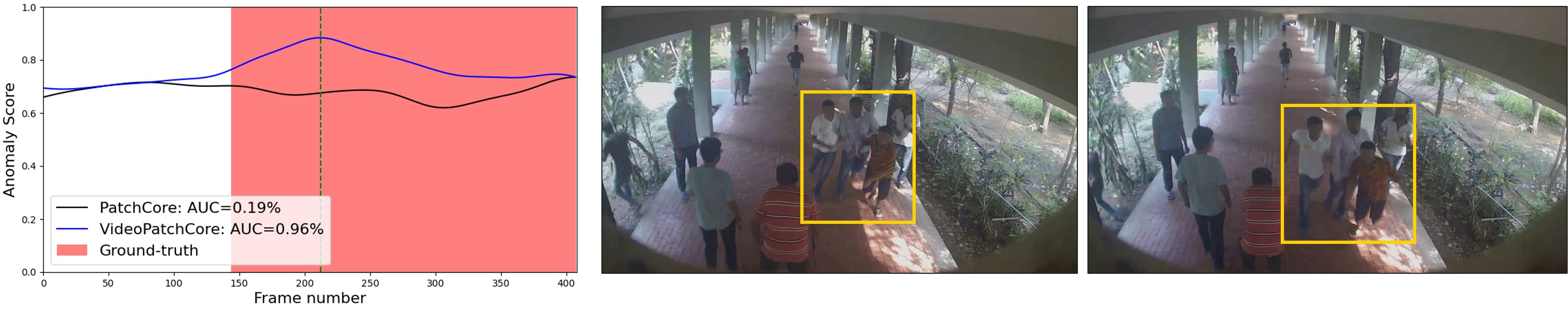}
        \caption{Sudden running}
        \label{fig:sfigure2a}
    \end{subfigure}
    \begin{subfigure}[h]{1.0\linewidth}
        \includegraphics[width=\textwidth]{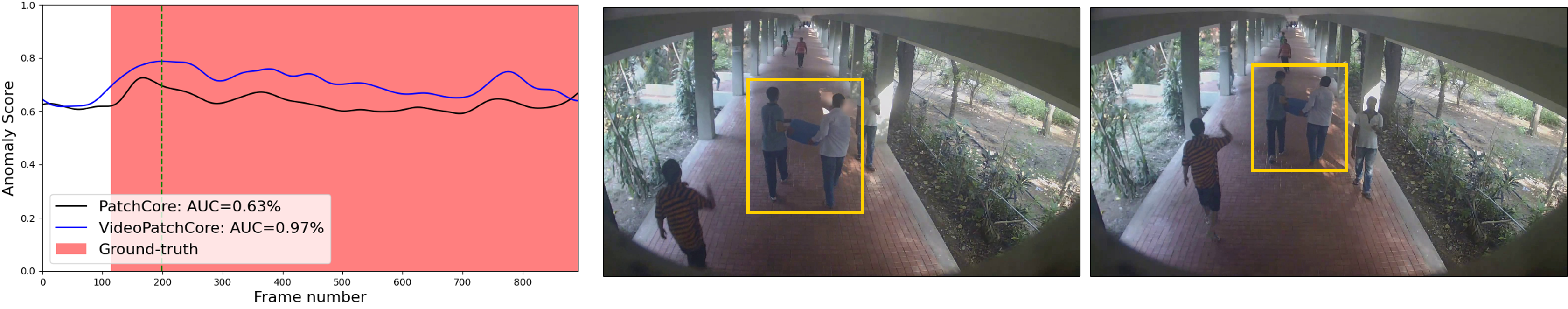}
        \caption{Suspicious object}
        \label{fig:sfigure2b}
    \end{subfigure}
    \caption{Comparison of anomaly scores between VPC and PC in the Corridor dataset.}
    \label{fig:sfigure2}
\end{figure}
\cref{fig:sfigure2a} shows anomaly frames depicting the action of sudden running, necessitating consideration of motion information. Therefore, temporal memory plays a crucial role in this scenario. \cref{fig:sfigure2b} shows anomaly frames depicting the action of moving a suspicious object, necessitating consideration of relationship between the person and object. Therefore, high-level semantic memory plays a crucial role in this scenario. VPC, which employs high-level and temporal memory, effectively distinguishes between abnormal and normal frames. In contrast, PC tends to produce false positives.
\subsection{Object-wise Anomaly Score Visualization}

\begin{figure}
    \centering
    \begin{subfigure}[h]{1.0\linewidth}
        \centering
        \includegraphics[width=\textwidth]{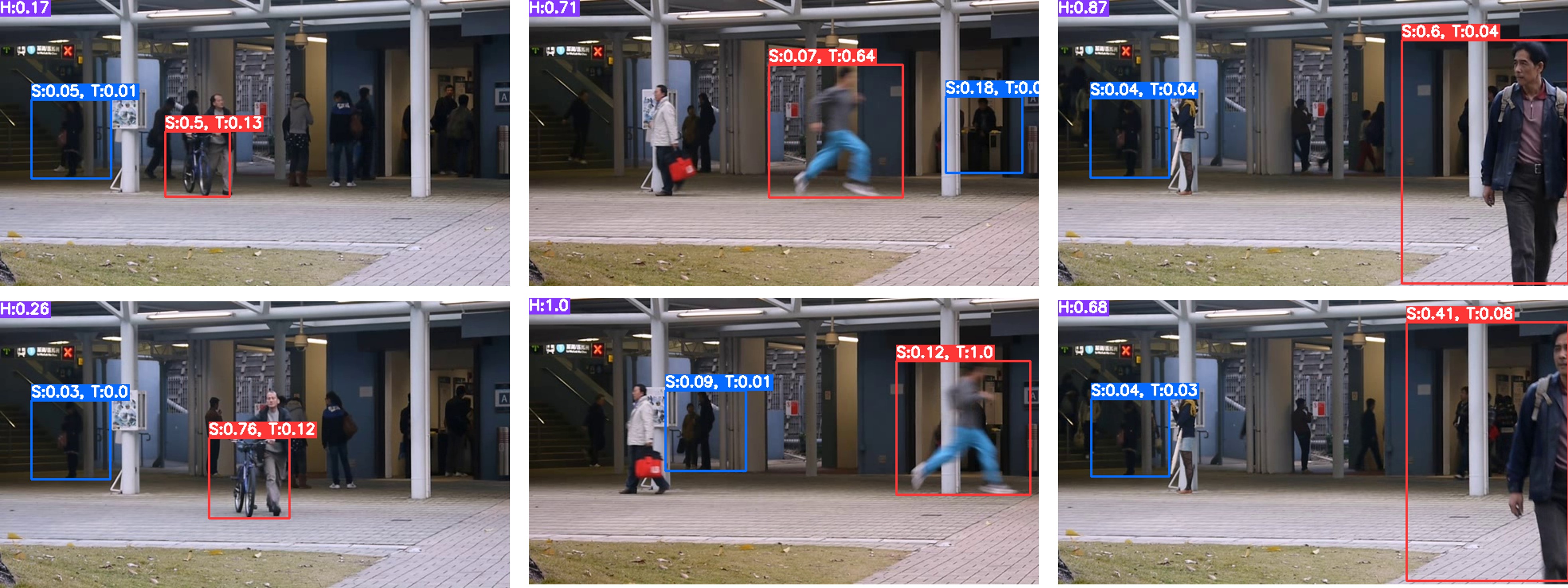}
        \caption{Avenue \emph{(Bicycle, Running, Wrong direction)}}
        \label{fig:sfigure3a}
    \end{subfigure}
    \begin{subfigure}[h]{1.0\linewidth}
        \includegraphics[width=\textwidth]{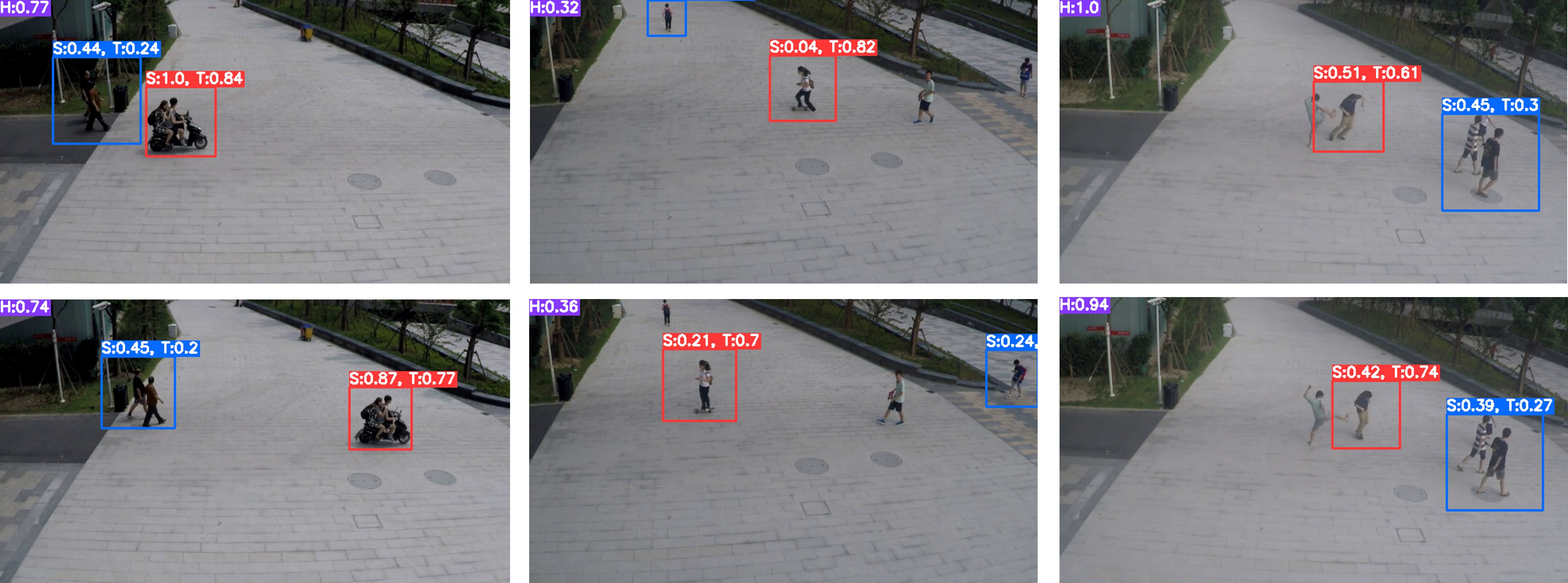}
        \caption{SHTech \emph{(Motorcycle, Skateboarding, Fighting)}}
        \label{fig:sfigure3b}
    \end{subfigure}
    \begin{subfigure}[h]{1.0\linewidth}
        \includegraphics[width=\textwidth]{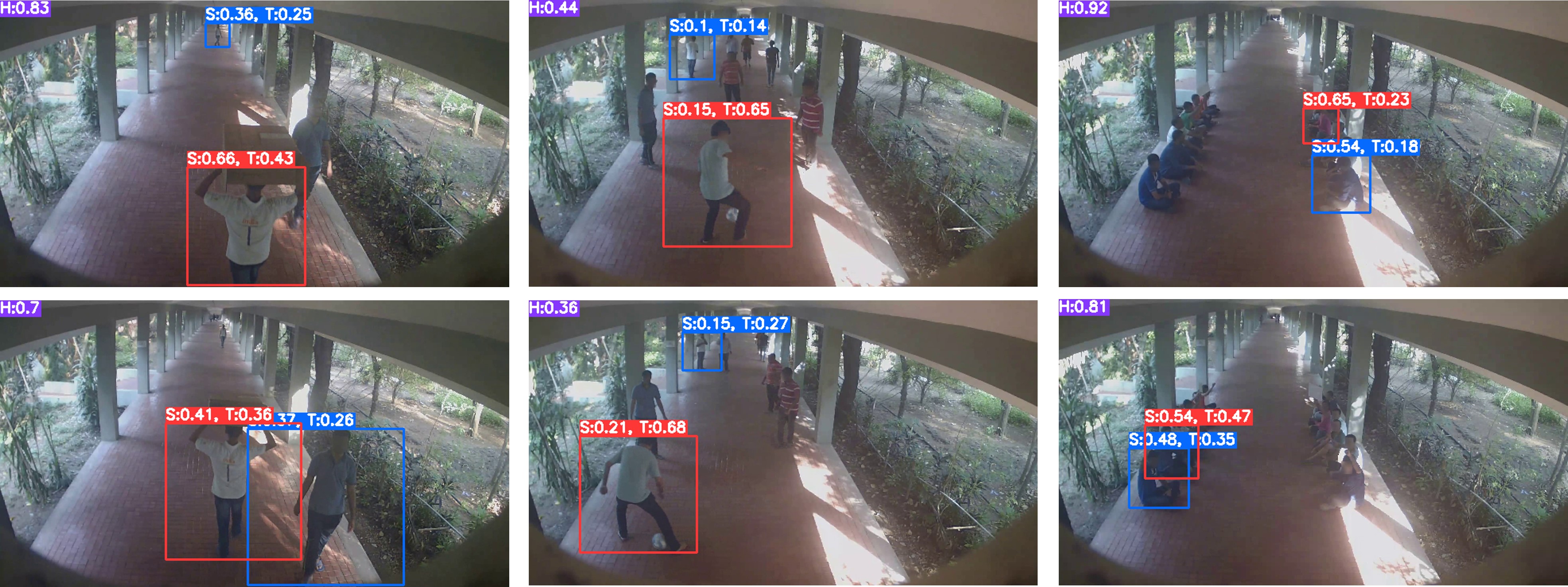}
        \caption{Corridor \emph{(Suspicious object, Playing with ball, Protest)}}
        \label{fig:sfigure3c}
    \end{subfigure}
    \caption{Three anomaly scores per object on the Avenue, SHTech, and Corridor datasets. H: high-level anomaly score, S: spatial anomaly score, T: temporal anomaly score.}
    \label{fig:sfigure3}
\end{figure}

For a deeper understanding of memory effectiveness, we computed anomaly scores using each memory module for two specific objects shown in \cref{fig:sfigure3}. One is the most anomalous object, and the other is the most normal object, as determined by the proposed model. Each object is characterized by the spatial (S) and temporal anomaly scores (T), while frames containing these objects are assigned a high-level anomaly score (H). The experimental results demonstrate precise prediction of anomalous objects within frames by the proposed model, with each memory module effectively fulfilling its role in various scenarios. For instance, in the bicycle scenario involving abnormal appearance, S increases due to spatial memory, while in the running scenario with abnormal behavior, T increases due to temporal memory. Finally, challenging anomalies such as "wrong direction" in the local stream are detected by H increasing in high-level memory. This validates the effectiveness of the three memory modules in VAD.

\begin{figure}
    \centering
    \begin{subfigure}[h]{0.9\linewidth}
        \centering
        \includegraphics[width=\textwidth]{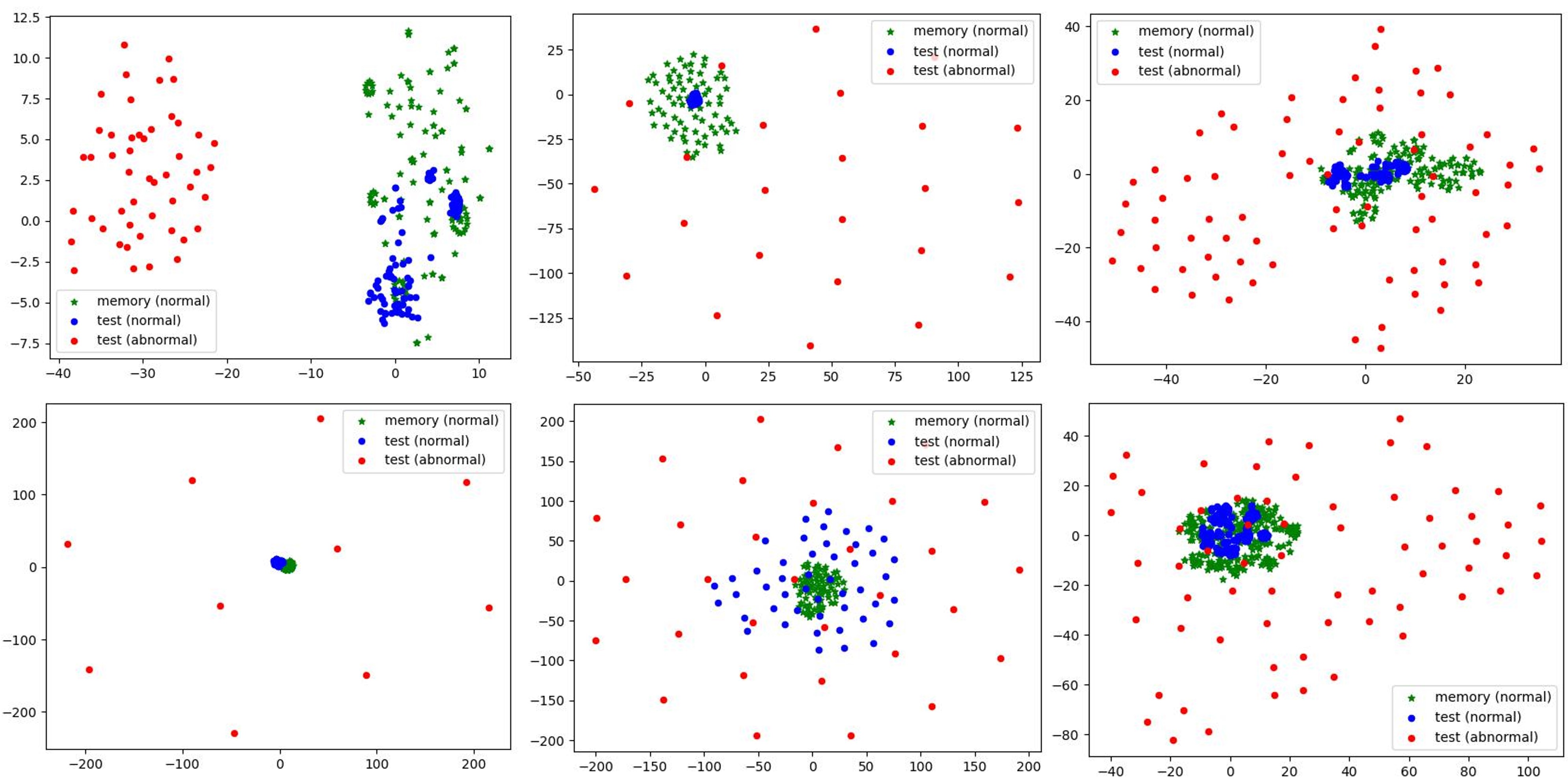}
        \caption{Spatial Memory \emph{(Avenue:01,03), (SHTech:18,28), (Corridor:09,35)}}
        \label{fig:sfigure4a}
    \end{subfigure}
    \begin{subfigure}[h]{0.9\linewidth}
        \includegraphics[width=\textwidth]{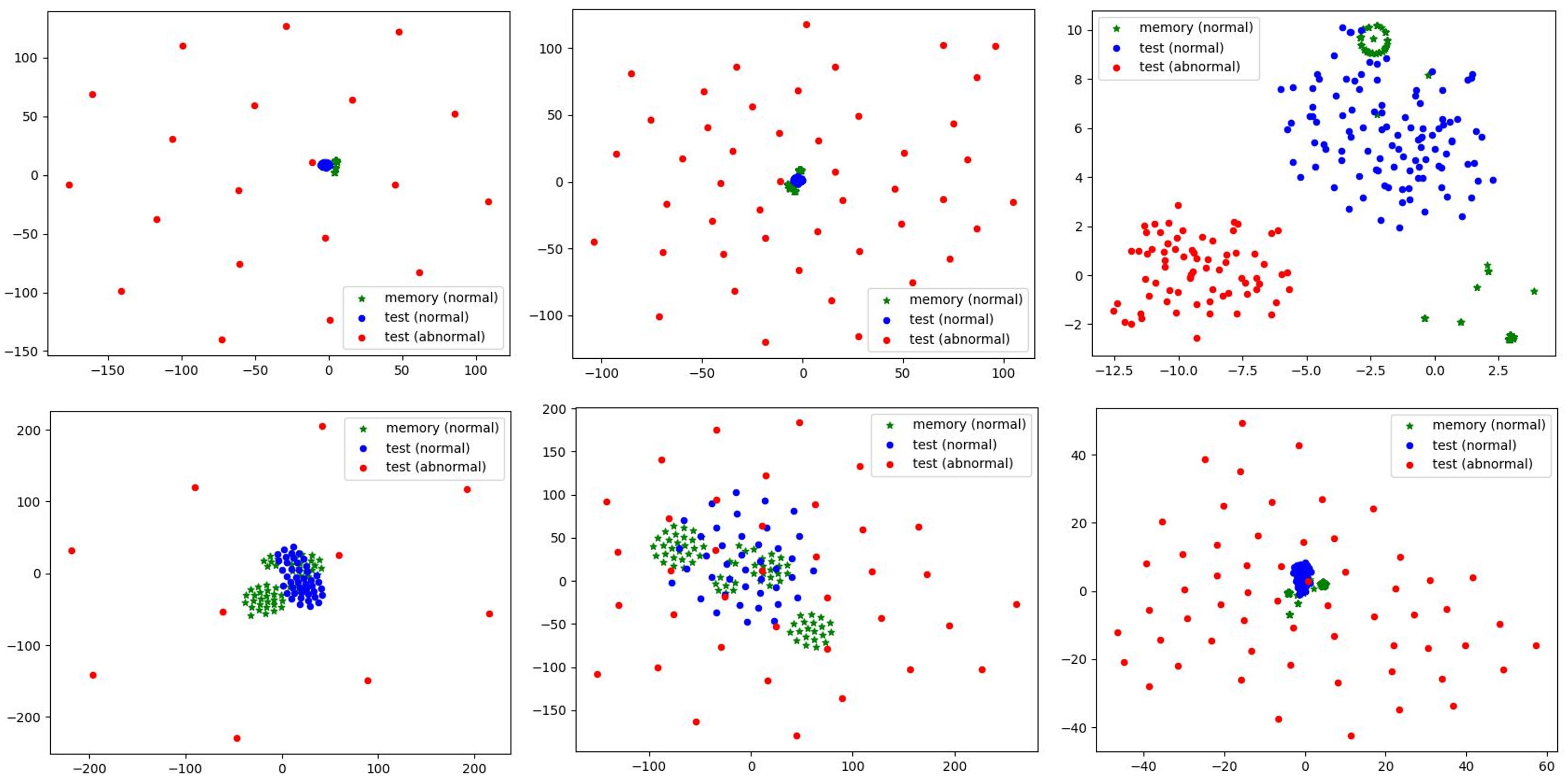}
        \caption{Temporal Memory \emph{(Avenue:10,13), (SHTech:24,86), (Corridor:45,140)}}
        \label{fig:sfigure4b}
    \end{subfigure}
    \begin{subfigure}[h]{0.9\linewidth}
        \includegraphics[width=\textwidth]{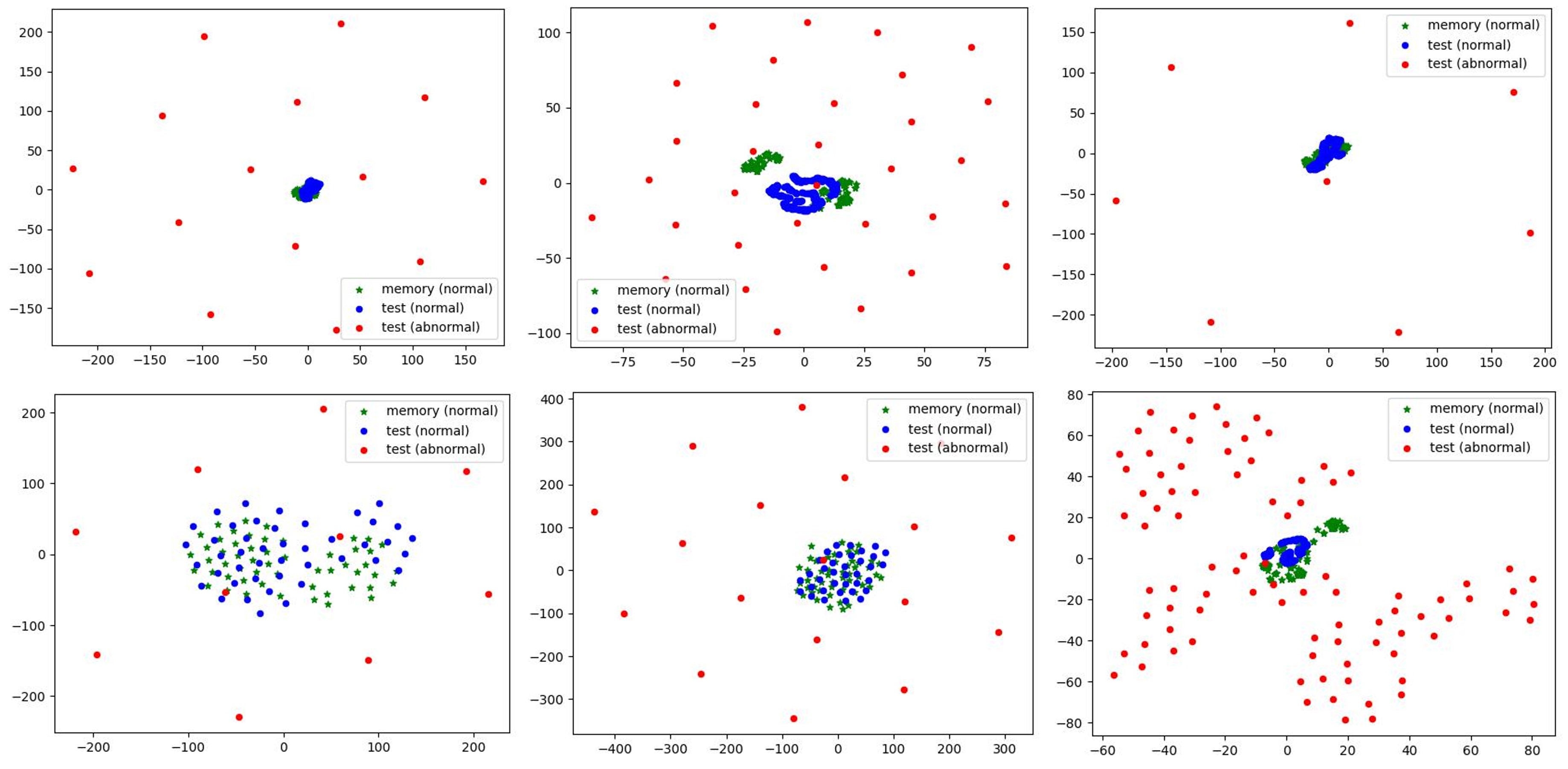}
        \caption{High-level Memory \emph{(Avenue:02,13), (SHTech:49,68), (Corridor:18,85)}}
        \label{fig:sfigure4c}
    \end{subfigure}
    \caption{t-SNE visualization of memory and test patches. The columns, from left to right, correspond to the Avenue, SHTech, and Corridor datasets.}
    \label{fig:sfigure4}
\end{figure}

\subsection{Patch Visualization}

\cref{fig:sfigure4} depicts the t-SNE plots of normal patches stored in memory (denoted by green '*'), along with normal (blue 'o') and anomalous patches (red 'o') from the test data. The results show that normal patches are clustered closely together compared to anomalous patches, aligning closely with the distribution of memory patches. In contrast, anomalous patches exhibit a wider dispersion and tend to be farther away from the memory patches. These findings suggest that the proposed memory effectively stores the normalcy of videos, making it suitable for VAD.

\begin{table}[t]
\centering
\caption{Comparison of AUROC scores for all memory banks on the Avenue, SHTech 
and Corridor datasets. The best results are \textcolor{red}{red} and the second best results are \textcolor{blue}{blue}.}
\label{tab:detailed_auc}
\begin{tabular}{c|cccccc}
\toprule
\textbf{\quad Avenue \quad} & \textbf{\quad 1\% \quad} & \textbf{\quad 10\% \quad} & \textbf{\quad 25\% \quad} & \textbf{\quad 50\% \quad} & \textbf{\quad 75\% \quad} & \textbf{\quad 99\% \quad} \\
\hline\hline
\addlinespace[2pt]
\textbf{Spatial} & \textcolor{red}{0.848} & \textcolor{blue}{0.831} & 0.828 & 0.825 & 0.828 & 0.828 \\
\textbf{Temporal} & \textcolor{red}{0.669} & \textcolor{red}{0.669} & \textcolor{red}{0.669} & \textcolor{red}{0.669} & \textcolor{red}{0.669} & \textcolor{red}{0.669} \\
\textbf{High-level} & 0.844 & \textcolor{blue}{0.845} & \textcolor{red}{0.848} & 0.844 & 0.844 & 0.844 \\
\textbf{Total} & \textcolor{red}{0.928} & \textcolor{blue}{0.918} & 0.914 & 0.912 & 0.912 & 0.912 \\
\midrule
\end{tabular}

\begin{tabular}{c|cccccc}
\midrule
\textbf{\quad SHTech \quad} & \textbf{\quad 1\% \quad} & \textbf{\quad 10\% \quad} & \textbf{\quad 25\% \quad} & \textbf{\quad 50\% \quad} & \textbf{\quad 75\% \quad} & \textbf{\quad 99\% \quad} \\
\hline\hline
\addlinespace[2pt]
\textbf{Spatial} & \textcolor{red}{0.748} & 0.744 & \textcolor{blue}{0.747} & \textcolor{blue}{0.747} & 0.746 & 0.746 \\
\textbf{Temporal} & \textcolor{red}{0.788} & \textcolor{red}{0.788} & \textcolor{red}{0.788} & \textcolor{red}{0.788} & \textcolor{red}{0.788} & \textcolor{red}{0.788} \\
\textbf{High-level} & 0.671 & \textcolor{blue}{0.675} & \textcolor{red}{0.684} & 0.673 & 0.673 & 0.674 \\
\textbf{Total} & 0.846 & \textcolor{blue}{0.850} & \textcolor{red}{0.851} & \textcolor{red}{0.851} & \textcolor{red}{0.851} & \textcolor{red}{0.851} \\
\midrule
\end{tabular}

\begin{tabular}{c|cccccc}
\midrule
\textbf{\quad Corridor \quad} & \textbf{\quad 1\% \quad} & \textbf{\quad 10\% \quad} & \textbf{\quad 25\% \quad} & \textbf{\quad 50\% \quad} & \textbf{\quad 75\% \quad} & \textbf{\quad 99\% \quad} \\
\hline\hline
\addlinespace[2pt]
\textbf{Spatial} & 0.690 & \textcolor{blue}{0.705} & \textcolor{blue}{0.705} & \textcolor{blue}{0.705} & \textcolor{red}{0.706} & \textcolor{blue}{0.705} \\
\textbf{Temporal} & \textcolor{red}{0.735} & \textcolor{red}{0.735} & \textcolor{red}{0.735} & \textcolor{red}{0.735} & \textcolor{red}{0.735} & \textcolor{red}{0.735} \\
\textbf{High-level} & 0.664 & 0.672 & 0.673 & \textcolor{blue}{0.674} & \textcolor{red}{0.675} & 0.660 \\
\textbf{Total} & 0.760 & \textcolor{red}{0.764} & \textcolor{blue}{0.763} & \textcolor{blue}{0.763} & \textcolor{blue}{0.763} & \textcolor{red}{0.764} \\
\bottomrule
\end{tabular}
\end{table}

\section{Quantitative Evaluation}

\subsection{Detailed Analysis of Subsampling Ratio}
We conducted a detailed analysis of performance changes based on subsampling ratios in the Avenue, Shanghai, and Corridor datasets as shown in \cref{tab:detailed_auc}. In the SHTech and Corridor datasets, the performance was better when using more memory, whereas in the Avenue dataset, the performance tended to be higher with less memory.

These results can be explained with two reasons. First, due to the diversity of normal data in the SHTech and Corridor datasets, the performance improves as more normal patches are stored in memory. Second, the Avenue dataset contains many action anomalies, such as easy running, making it challenging to distinguish them from normal instances without using temporal information. Therefore, from a spatial memory perspective, filtering out normal patches that resemble anomalies enhances the performance. Meanwhile, other memories that utilize temporal information exhibit consistent performance regardless of their size. Consequently, using only 1\% of memory overall yields the best performance.

However, as evidenced by the experimental results, the performance difference between using 10\% and 99\% of memory is very small. Therefore, in practical use, sufficiently good performance can be maintained even with memory usage set at 10\% or lower.
\end{document}